\documentclass[conference]{IEEEtran}
\usepackage{cite}
\usepackage[utf8]{inputenc} 
\usepackage[T1]{fontenc}    
\usepackage{url}            
\usepackage{booktabs}       
\usepackage{amsfonts}       
\usepackage{nicefrac}       
\usepackage{microtype}      
\usepackage{graphicx}
\usepackage{subfigure}
\usepackage{caption} 
\usepackage{amsmath}
\usepackage{algorithm}  
\usepackage[noend]{algpseudocode} 

\IEEEoverridecommandlockouts
\begin{document}
\captionsetup[figure]{labelfont={},name={Fig.},labelsep=period}
\captionsetup[table]{labelfont={},name={TABLE.},labelsep=period}
\title{UCT-ADP Progressive Bias Algorithm for Solving Gomoku}

\renewcommand{\thefootnote}{\fnsymbol{footnote}}

\author{\IEEEauthorblockN{Xu Cao and Yanghao Lin}
\IEEEauthorblockA{School of Data Science\\
Fudan University\\
Shanghai, China\\
Email: irohcao@gmail.com; chronoslyh@gmail.com}
}


%


\maketitle

\begin{abstract}
  We combine Adaptive Dynamic Programming (ADP), a reinforcement learning method and UCB applied to trees (UCT) algorithm with a more powerful heuristic function based on Progressive Bias method and two pruning strategies for a traditional board game Gomoku. For the Adaptive Dynamic Programming part, we train a shallow forward neural network to give a quick evaluation of Gomoku board situations. UCT is a general approach in MCTS as a tree policy. Our framework use UCT to balance the exploration and exploitation of Gomoku game trees while we also apply powerful pruning strategies and heuristic function to re-select the available 2-adjacent grids of the state and use ADP instead of simulation to give estimated values of expanded nodes. Experiment result shows that this method can eliminate the search depth defect of the simulation process and converge to the correct value faster than single UCT. This approach can be applied to design new Gomoku AI and solve other Gomoku-like board game.
\end{abstract}

\

\begin{IEEEkeywords}
\emph{adaptive dynamic programming; monte carlo tree search; gomoku; exponential heuristic; progressive bias.}
\end{IEEEkeywords}

%
\IEEEpeerreviewmaketitle

\

\section{Introduction}
  Gomoku is a traditional strategical board game designed for two players playing on the 15x15 grids Go board or other similar board. The players are alternatively putting their own signs (cross and circle or black and white Go chessman) on the board until one of them first manages a continuous line of five or more his or her same signs in the vertical, horizontal or diagonal direction and thus becomes the winner. This is the freestyle rules of Gomoku.
\\  
\par
  To solve the freestyle Gomoku, we design a new exponential reward increasing heuristic function and combine it with these classic algorithms and prior methods — UCB applied to Monte-Carlo Tree Search, Adaptive Dynamic Programming, Victory of Continuous Four (VCF) and Victory of Continuous Three (VCT) pruning strategy. The Monte-Carlo Tree Search (MCTS) is one of the most significant and famous search algorithms for strategical board game. In the case of Go , it is proved better than any other traditional tree search algorithms such as Minimax Tree Search with Alpha-Beta Pruning. However, MCTS still have some drawbacks which restrict its search ability. Obviously, One of the most significant problems is the time and space complexity increasing exponentially with the simulation search depth \cite{A Survey of Monte Carlo Tree Search Methods}. If MCTS’s simulation depth choice and random sampling strategies are improper, it will waste a great amount of time in the shallow depth of MCTS’s search tree structure thus can not explore winning or defeat subtrees. To handle this problem, we propose a shallow forward neural network evaluation function trained by Adaptive Dynamic Programming (ADP), a Progressive Bias-UCB based heuristic strategy and two pruning strategy called VCF and VCT to improve the search efficiency of Gomoku game trees.
\\ 
\par
  Next section provides a brief description of the MCTS and ADP research related work, then section 3 briefly introduces the MCTS algorithm based on UCB. Section 4 presents our new UCT-ADP Progressive Bias Gomoku solver. Section 5 presents and discusses the experimental and competition results among our AI and other Gomoku AI solvers, while it also analyses the convergence of our Gomoku solver. In the final section, we summarize the whole paper and offered some possible improvement direction for future research.
\\
\par

\section{Related Work}
  Monte Carlo Tree Search (MCTS) is a classic search algorithm for obtaining optimal decisions by randomized self-play simulation in the adapt game regions and building a search tree architecture according to the results. Due to its spectacular success in board game Go, It started to be used in many other Go-like board games such as Gomoku and Reversi. Browne summarized the results from the key game to which MCTS methods had been applied and introduced a series of improvement strategies \cite{A Survey of Monte Carlo Tree Search Methods}. One of the most important improvement strategy methods is Upper Confidence Bounds for Trees (UCT): MCTS with a UCB tree selection policy \cite{A Survey of Monte Carlo Tree Search Methods}\cite{Monte Carlo Tree Search Method for AI Games}. Although MCTS was alreadly applied to solving Gomoku, it did not work better than conventional Alpha-Beta Pruning algorithm based on VCF and VCT heuristic function in gomocup competition. Kang proposed to use Progressive Bias, a combination between UCB and Gomoku heuristic based strategy to increase the accuracy and save computational time in selection process \cite{Effective Monte-Carlo tree search strategies for Gomoku AI}. Wang and Mohandas applied Genetic Algorithms (GA) to optimize the game tree search space of Gomoku and obtain great effect \cite{Evolving Gomoku solver by genetic algorithm}\cite{A.I for Games with High Branching Factor}. Besides, most of the available improvement that suits Go can also apply to Gomoku, such as UCT–RAVE \cite{Monte-Carlo Tree Search and Rapid Action Value Estimation in Computer Go}, SO-ISMCTS and MO-ISMCTS \cite{Information Set Monte Carlo Tree Search}. All these modified algorithms are proved to perform well in the gomocup contest held every year. 
\\
\par
  Except the MCTS method, Adaptive Dynamic Programing (ADP) can also be used to design Gomoku solver. ADP is a approach based on reinforcement learning which does not need information about the Markov decision process (MDP). This method concerns how agent would like to take actions in an unknown environment so as to explore states and obtain maximized  rewards \cite{Adaptive game AI for Gomoku}\cite{Reinforcement Learning for Build-Order Production in StarCraft II}. Zhao first applied Adaptive Dynamic Programing (ADP) based on neural network to train gomoku AI model by playing against itself \cite{Self-teaching adaptive dynamic programming for Gomoku}. Then, Tang combined MCTS with ADP by the weighted sum of ADP and its corresponding winning probability of MCTS \cite{ADP with MCTS algorithm for Gomoku}. They observed that UCT-ADP mixed model is able to get higher benefits than both UCT-MCTS and ADP alone. In a word, ADP is a very effective and powerful winning rate evaluation model which can realizes a human-like intuitive analysis of the Gomoku-board situation, but it cannot analyze the success rate after multiple steps in the future.
\\
\par
  Inspired by Zhao and Tangs' research, we combine Monte Carlo Tree Search with ADP into Gomoku in another novel way, as well as improving its architecture and strategy by a more complex and effective heuristic model than a single weighted sum model. Accordingly, we actually obtain a stronger Gomoku-AI that can defeat MCTS, ADP, and also single weighted sum UCT-ADP agent.
\\
\par

\section{Monte Carlo Tree Search with UCB}
  Monte Carlo tree search (MCTS) is a tree search algorithm based on tree architecture, which can be considered as a sort of tradeoff policy between exploration and utilization, and it can explore more enormous board grid space than any other tree search algorithm \cite{A Survey of Monte Carlo Tree Search Methods}. Some papers believe that MCTS can be a best-first search by precessing a large number of random simulation \cite{Effective Monte-Carlo tree search strategies for Gomoku AI} and thus converge to the best solution. Therefore, it obtains high accuracy with random sampling and decreases time cost that spends for calculating evaluation values. 
\\
\par
  The 4 procedural steps of MCTS is shown in Fig.1, including Selection, Expansion, Simulation, and Back-propagation. The first step is Selection. The aim of selection is to choose one of the best nodes worth exploring in the tree. In Gomoku, the general strategy is to select from the 1-grid or 2-grid adjacent child nodes which are empty. After determining the range for selection, we need to choose a suitable policy to judge the selecting priority order. In general, Upper Confidential policy is the core selection process of MCTS, which makes proper selection strategy by Upper Confidential Bound (UCB). Other pruning and heuristic methods can be combined and improved along the basic idea of UCB.  UCB is a well-known algorithm applying to solve the multi-armed bandit problem. As formula (\ref{UCB formula}) shows, UCB can trade-off exploration and exploitation. In this function, N is the total nodes amount, and $n_i$ is the child nodes number belong to node $i$. $v_i$ is the value that the total winning rate of node that has all results of child nodes. $k$ is a constant value determining the trade-off between exploration and utilization. In our experiment, we set $k$ to $\sqrt{2}$.

\begin{equation}
\begin{aligned}
UCB=v_i+k*\sqrt{\frac{ln(N)}{n_i}}
\label{UCB formula}
\end{aligned}
\end{equation}
\par
  Formula (\ref{UCB formula}) can help AI select the child node which has the largest UCB value. Then at the Expansion step, the algorithm will create a new child node of the selected node. The third step is Simulation, which plays out a game from the new expanded child node until arriving at an outcome of Win (+1) or Lose (+0). It is worth mentioning that in the case of Gomoku game, even only using 2-adjacent grid selection, both the breadth and the depth of the simulation tree can be unacceptably large. This is because when exploring deeper nodes, the number of possible paths will increase exponentially. Based on the idea that the simulation process essentially serves as an evaluation function of non-terminal states, we replace the time-consuming simulation process with a winning rate evaluation function trained by ADP (Fig. \ref{UCT-ADP process}), which will be explained in the next section. The fourth step is Back-propagation, which is to back propagate the score of the node from the previous expansion to all the previous parent nodes, and update the winning values (0 and 1 in conventional UCT strategy) and visit times of these nodes to facilitate the calculation of the UCB value. All of the values are stored in a tree structure as a global variable.

\begin{figure}[!tp]
  \centering
  \includegraphics[width=1.\linewidth]{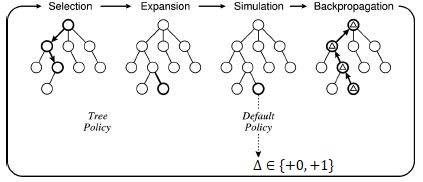} 
  \caption{One iteration of the general MCTS approach \cite{A Survey of Monte Carlo Tree Search Methods}.}
  \label{MCTS process}
\end{figure}

\begin{figure}[!tp]
  \centering
  \includegraphics[width=1.\linewidth]{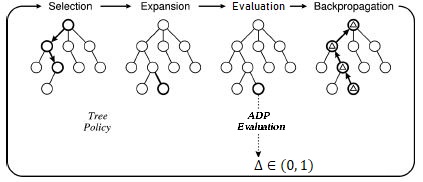} 
  \caption{One iteration of the UCT-ADP approach.}
  \label{UCT-ADP process}
\end{figure}

\par
  In brief, UCT can be considered as the situation that MCTS + UCB. Computing the UCB value in each selection period can help the UCT algorithm find out the balanced leaf nodes earlier than simple HMCTS algorithm, thus, it is faster and more effective than fully random MCTS such as HMCTS \cite{ADP with MCTS algorithm for Gomoku}.
\\
\par

\section{UCT-ADP Progressive Bias Algorithm}
  The first UCT-ADP algorithm designed by Tang uses individual results from MCTS and ADP respectively to calculate the weighted sum of winning rates to select the next Gomoku sign position \cite{ADP with MCTS algorithm for Gomoku}. Although this method indeed improves the search results, it does not highly improve the speed of MCTS simulations and final convergency. We have redesigned a new algorithm that uses progressive bias to combine UCB with a priori exponential heuristic function and also reconstruct the UCT tree structure. The progressive bias is able to realize select, expand and simulate until a certain depth. Then at back-propagation stage, if there is no winner in this certain depth, it will return and update the final winning rate calculated by ADP. Our algorithm make full use of the characteristics of ADP's board situation evaluation, and obtain better results than weighted summation UCT-ADP. This section will focus on four main parts of our new designed algorithm.

\subsection{Exponential Heuristic Function}
  To reduce most of the computational cost in selection process, it is significant to design an efficient heuristic algorithm which is simple and has practical form. The heuristic function needs to be able to evaluate the rewards of the next step on the board, including the offense rewards and the defense rewards. Since Gomoku is not a complicated game, it is not difficult to build such a heuristic function. Jun Hwan Kang \cite{Effective Monte-Carlo tree search strategies for Gomoku AI} proposed using formula (\ref{heuristic formula1}) as the Gomoku heuristic function.

\begin{equation}
\begin{aligned}
H_i=\sum{\{(L_{open}^2+(\frac{L_{hclose}}{2})^2\}}
\label{heuristic formula1}
\end{aligned}
\end{equation}

\par
\begin{figure}[!tp]
  \centering
  \includegraphics[width=1.\linewidth]{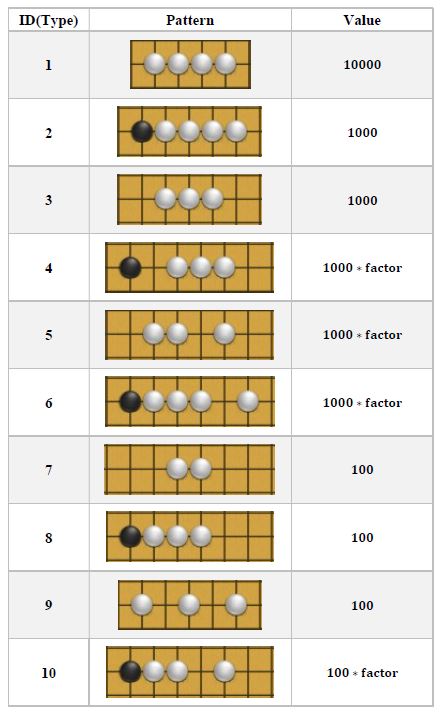} 
  \caption{Example of the patterns and their heuristic value.}
  \label{pattern heuristics}
\end{figure}

  Variable $L_{open}$ is length of line that has no opponent's chessman at two terminals, while variable $L_{hclose}$ is length of line that has only one opponent's chessman at its terminals. However, this heuristic function has a bit of problem. First, it ignores the potential for discontinuous offense patterns, such as the one known as ‘jump three’(Pattern ID 5 in Fig.  \ref{pattern heuristics}). Second, its valuation of offense is unreasonable. In freestyle rules of Gomoku, ‘live three’ ($L_{open}=3$,Pattern ID 3 in Fig. \ref{pattern heuristics}) and ‘sleep four’ ($L_{hclose}=4$,Pattern ID 2 in Fig. \ref{pattern heuristics}) should have similar offense effect, but they differ too much in this heuristic (9 and 4).
\\
\par
  To solve this problem, we design a new sort of heuristic function: exponential heuristic function. It can consider more states and situations than Jun Hwan Kang’s simple heuristic and can also be used to involve in pruning process.

\begin{equation}
\begin{aligned}
H_i=\sum{\{10^{L_{open}}*factor^j+10^{L_{hclose}-1}*factor^k\}}
\label{heuristic formula2}
\end{aligned}
\end{equation}

  Variable $L_{open}$ is length of line that has no opponent's chessman at two terminals and some similar situation with a decay factor, while variable $L_{hclose}$ is length of line that has only one opponent's chessman at its terminals and some similar situation with a decay factor. Whether the pattern's heuristic needs to multiply the decay factor is determined according to its a priori attack reward, as shown in Fig. \ref{pattern heuristics}. In our experiment, the decay factor is a constant. We set it to 0.90.
\\
\par
  Using this exponential formula we obtain 32 patterns’ heuristic values. Fig. \ref{pattern heuristics} below shows parts of it. The full heuristic function and patterns are in our codes. Meantime, we also use these patterns to train the neural network of Adaptive Dynamic Programming.
\\

  The Progressive Bias is a UCB plus heuristic based strategy that can be more time-expensive but also more accurate than UCB to select moves.  The aim of progressive bias is to get bias in selection phase to add more significant nodes to a pre-selected list with priori heuristic knowledge. The formula (\ref{PB formula}) shows our Progressive Bias-UCB formula. It consists of basic form of UCB and a exponential heuristic values gain. 

\begin{equation}
\begin{aligned}
UCB=v_i+k_1*\sqrt{\frac{ln(N)}{n_i}}+ k_2*\frac{H_{i}}{max(H)}
\label{PB formula}
\end{aligned}
\end{equation}

  In formula (\ref{PB formula}),$H_{i}$ is the heuristic value that we calculate in formula (\ref{heuristic formula2}). $max(H)$ is the highest heuristic value (100000 for five row win state) or can be set as a changeable hyperparameter. $k_1$ 's meaning is same to $k$ in formula (\ref{UCB formula}), while $k_2$ is a constant which can set the importance of the exponential heuristic values gain. In a word, the exponential heuristic function is a a simple mathematical model which can both distinguish different levels of rewards and punishments to a certain extent, and do not need numerous calculation time.

\subsection{Selecting Moves Pre-select Function}
\begin{figure}[!tp]
\centering
\subfigure(a)
{
	\centering          
	\includegraphics[width=0.4\linewidth]{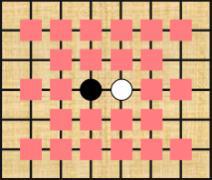}   
}
\subfigure(b)
{
	\centering     
	\includegraphics[width=0.3475\linewidth]{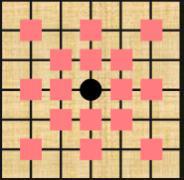}  
}
\caption{Example of the 2-adjacent legal grid moves \cite{An algorithmic design and implementation of outer-open gomoku}.}
\label{adjacents}
\end{figure}

  Due to less restrictions in free-style Gomoku, the number of legal moves is pretty large. If we choose all legal moves in the board, it will be difficult for the selection and simulation process in MCTS in a short time. Therefore we need use suitable adjacent grids moves to replace all legal moves. According to our designed heuristic function, the available point should be the 2-adjacent grids as Fig. \ref{adjacents} shows. Supposed the available is 1-adjacent grids, it will miss some significant points such as the right Go chessman in pattern ID 5 in Fig. \ref{adjacents}. Choosing 2-adjacent grids is proved to be an effective way to reduce the branching factor of Gomoku \cite{An algorithmic design and implementation of outer-open gomoku}.
\\
\par

\subsection{Victory of Continuous Four and Victory of Continuous Three}
  The Victory of Continuous Four (VCF) / Victory of Continuous Three (VCT) search are the terms of traditional Gomoku and does not seem to be understood easily by layman. Each of them can be considered as very significant strategies for alpha-beta pruning. The VCF refers to continuously manufacturing a sleep four attack like pattern ID 2 in Fig. \ref{pattern heuristics} in the case of the opponent do not have opportunity to fight back, until obtain the final  winning state or can not generate new VCF situation. VCT refers to continuously manufacturing a alive three attack like pattern ID 3 or ID 5 in Fig. \ref{pattern heuristics} in the case of the opponent do not have opportunity to fight back, until arriving at the final five or can not generate any VCF or VCT circumstance. 
\\
\par
  Therefore, we need use human designed exponential heuristics function to detect VCF and VCT. When the heuristic function value is on the order of $10^3$ or more, the attack is an effective attack that conforms to VCF and VCT. This feature of the heuristic function can be used for pruning in UCT's selection stage. First, the Gomoku agent should detect whether there is a victory situation for enemy or whether it is possible to carry out a counterattack. For example, if the enemy's next step can also lead to his or her continuous four like pattern ID 1, in this case, alive three is not a effective attack choice for us while we have to choose sleep four attack or try to hinder the enemy's continuous four attack in the future. In summary,  pruning method should conform to the direction of right decision and on the other hand, it is able to decrease the potential selection grids enormously. In our algorithm, we calculate the heuristic value for all of our select moves. If the value or situation match VCF or VCT, the corresponding grid will append to a preferred list.
\\

\subsection{Adaptive Dynamic Programming(ADP)}
\begin{figure}[!tp]
  \centering
  \includegraphics[width=1.0\linewidth]{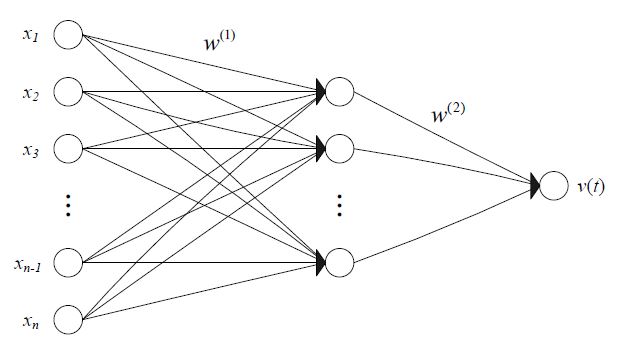}  
  \caption{The structure of the neutral network \cite{Self-teaching adaptive dynamic programming for Gomoku}.}
  \label{MLP}
\end{figure}

  To evaluate Gomoku board winning rate situations, we utilize a Multi-Layer Perceptron (MLP) of 3 layers (Fig. \ref{MLP}). The input features of the MLP includes the number of the 32 patterns mentioned above and a one hot vector indicating who is to move next. The number of patterns are encoded in a specific way, where a number is represented by a vector of size 5 (Table \ref{adp inputs}). The activation function of the MLP is the sigmoid function as formula (\ref{sigmoid formula}) shows.

\begin{equation}
\begin{aligned}
g(x)=\frac{1}{1+e^{-x}}
\label{sigmoid formula}
\end{aligned}
\end{equation}

  Which normalizes the output value to the range (0,1). This allows us to interpret the output of the MLP as the winning probability of a player.

\par

  We train the MLP in an Adaptive Dynamic Programming (ADP) process where the agent plays Gomoku games against itself. The reward is set to 0 during the game. When a game ends, if the player wins, the reward is 1 \cite{Self-teaching adaptive dynamic programming for Gomoku}. Else if the opponent wins, the reward is 0. The prediction error is defined as:

\begin{equation}
\begin{aligned}
e(t)=\alpha[r(t+1)+\gamma{V}(t+1)-V(t)]
\end{aligned}
\end{equation}

\begin{table}[t]
  \caption{A\textsc{dp} I\textsc{nputs} \cite{Self-teaching adaptive dynamic programming for Gomoku}}
  \label{adp inputs}
  \centering
  \begin{tabular}{llllll}
    \cmidrule{1-6}
    Value of n     & input1  & input2  & input3  & input4  & input5 \\
    \midrule
    0  &0  &0  &0  &0  &0  \\
    1  &1  &0  &0  &0  &0  \\
    2  &1  &1  &0  &0  &0  \\
    3  &1  &1  &1  &0  &0  \\
    4  &1  &1  &1  &1  &0  \\
	$\textgreater$4  &1  &1  &1  &1  &(n-4)/2  \\
    \bottomrule
  \end{tabular}
\end{table}

where $\alpha$ is the learning rate and $\gamma$ is the discount factor (set to 1). $V$ is the output of the MLP. When the number of patterns is 32, it takes over 10000 games for the MLP to converge. 
\\
\par
The original MCTS uses simulation to receive an estimated value of the expanded nodes. Performing simulations in the Gomoku game domain can raise several issues. If we choose uniform random moves during the simulation process, information learned through one simulation will be limited due to the large branching factor and depth of Gomoku game trees. If we apply complicated heuristic knowledge, the computational cost per simulation will increase drastically, leading to a reduction in the number of simulations possible.
To deal with this dilemma, we use the MLP trained by ADP (referred to as ADP in the following section) to evaluate board situations instead of simulating play-outs to get winning (+1) or losing (+0) outcomes.

\begin{algorithm}[!tp]
	\caption{UCT-ADP Progressive Bias Algorithm}  
	\label{alg:Framwork}  
	\begin{algorithmic}  
		\Require  
		Root node $v_0$ with State $s_0$
		\Ensure  
		action $a$ corresponding to the highest value of UCT-ADP Progressive Bias
		\While {within computational budget}
		\State $v_l$ $\gets$ Tree Policy($v_0$);
		\State reward r $\gets$ Evaluate(s($v_l$));
		\State Back Update($v_l$, r);
		\EndWhile
		\\
		\Return action $a$(Best Child($v_0$))
		\State
		
		\Function{Tree Policy}{node v}
		\While{v is not in terminal state or v $\leq$ MSD}
		\If {v not fully expanded}
		\State
		\Return Expand(v);
		\Else
		\State $v$ $\gets$ Best Child($v$);
		\EndIf
		\EndWhile
		\\
		\Return $v$
		\EndFunction  
		\State
		
		\Function{Expand}{node v}
		\For{$a_i$ in v's action $a$}
		\If {$a_i$ can realize VCF or VCT strategy}
		\State add $a_i$ to A(s(v))
		\EndIf
		\EndFor
		\If{No action in A(s(v)}
		\State A(s(v)) $\gets$ all untried actions
		\EndIf
		\State choose random action a $\in$ from A(s(v));
		\State add a new child $v^{'}$ to v with s($v^{'}$) $\gets$ $f$(s(v), a) 
		\State and $a$($v^{'}$) $\gets$ $a$;
		\\
		\Return $v^{'}$
		\EndFunction
		\State
		
		\Function{Best Child}{node v,parameter c}
		\\
		\Return $argmax{(Q(v')/N(v')+c\sqrt{2lnN(v')/N(v')})}$
		\EndFunction
		\State
		
		\Function{Back Update}{node v, reward r)}
		\While{v is not null}
		\State{N(v) $\gets$ N(v)+1}
		\State{Q(v) $\gets$ Q(v)+r}
		\State{r $\gets$ 1-r}
		\State{v $\gets$ Parent(v)}
		\EndWhile
		\EndFunction
		
	\end{algorithmic}  
\end{algorithm}

\section{Experiments and Analysis}
\subsection{Convergence of UCT-ADP}
  The convergence of UCT to the minimax tree at a polynomial rate has been proved theoretically by L Kocsis et al \cite{Bandit based monte-carlo planning}. Replacing the simulation stage by ADP evaluation does not alter the conclusion as long as the evaluation gives correct value for ending game states (+0 for lose and +1 for win).
  \\
  \par
  To empirically demonstrate the workings of UCT-ADP in Gomoku games, we compare the convergence rate of UCT-ADP and UCT-ADP with progressive bias (UCT-ADP-PB) to its baseline models (weighted-sum\cite{ADP with MCTS algorithm for Gomoku}, UCT-DUMMY and UCT-SIMULATION) under several specific board situations. The UCT-DUMMY replaces the output of ADP with the constant 0.5 (indicating a 50\% win rate for every board situation). The UCT-SIMULATION randomly simulates a game to its end and receives a +1 or +0 value for back-propagation. The weighted-sum baseline model selects its move according to a weighted sum of ADP output and UCT-SIMULATION output. 
  \\
  \par
  We use failure rate \cite{Bandit based monte-carlo planning} to evaluate the performance of models, indicating the possibility of choosing non-optimal moves if stopped after a number of iterations. For each model, we sample its failure rate from 50 UCT trees. The failure rates are plotted as function of iterations (Fig. \ref{convergence1} and Fig. \ref{convergence2}). The faster the failure rate drops to 0, the better the model is. 
  \\
  \par
\begin{figure}[!tp]
    \centering
    \includegraphics[width=1.0\linewidth]{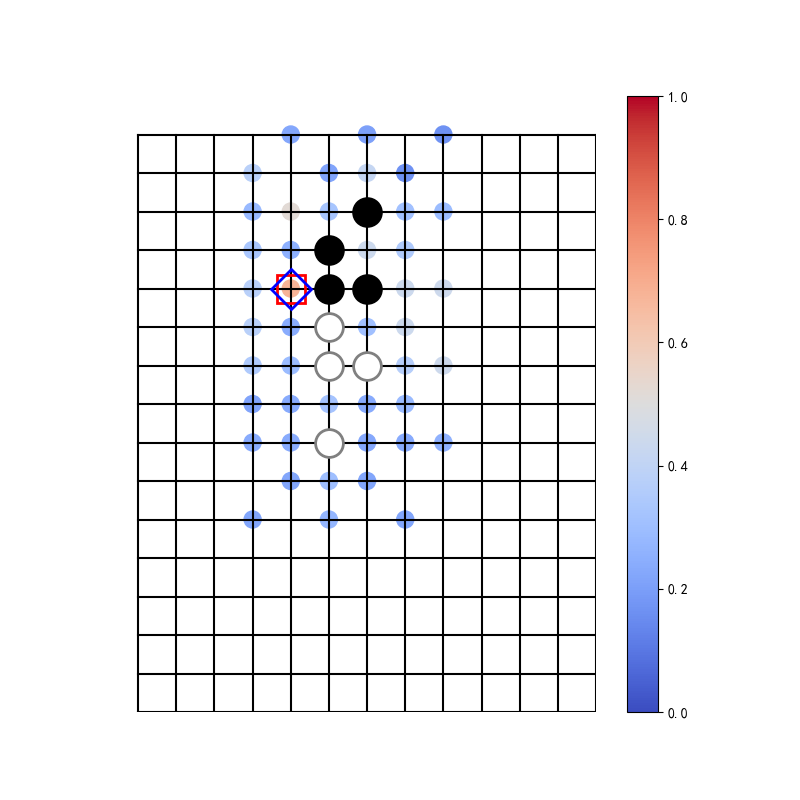}  
    \caption{Board Situation 1, Black's Turn (Predictions of ADP are plotted as colored circles, redder color indicates higher win rate for the black side. The best move predicted by ADP is marked with a red square. The true best move is marked with a blue diamond).}
    \label{board1}
  \end{figure}
\begin{figure}[h]
    \centering
    \includegraphics[width=1.0\linewidth]{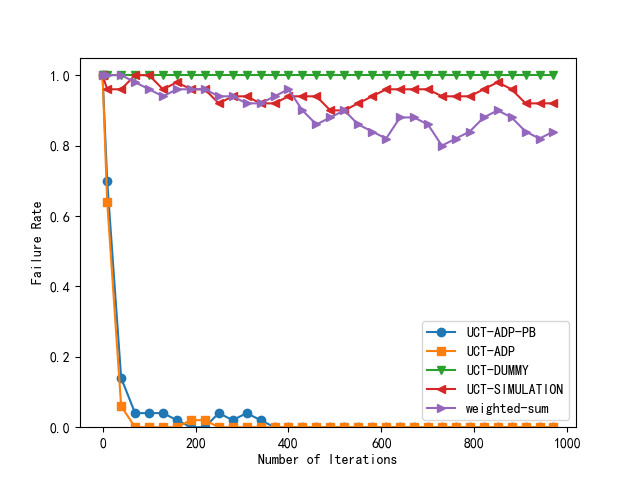}  
    \caption{Convergence to the correct move under Board Situation 1.}
    \label{convergence1}
  \end{figure}

  Under the board situation of Fig. \ref{board1}, the best move for the black side forms a double live-three which leads to the victory of black. ADP alone can already predict the best move, therefore speeding up the convergence of UCT. From Fig. \ref{convergence1}, we can see that UCT-ADP-PB and UCT-ADP converges to the correct move in 200 iterations while the baseline models seem to take much more iterations to converge. 
  \\
  \par
\begin{figure}[!tp]
    \centering
    \includegraphics[width=1.0\linewidth]{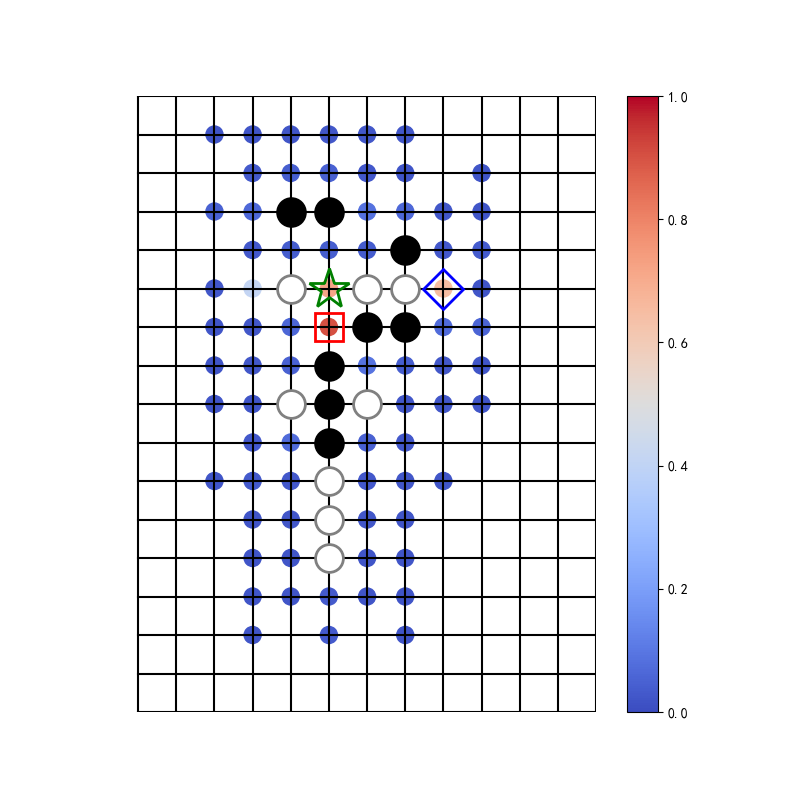}  
    \caption{Board Situation 2, Black's Turn (Predictions of ADP are plotted as colored circles, redder color indicates higher win rate for the black side. The best move predicted by ADP is marked with a red square. The true best move is marked with a blue diamond).}
    \label{board2}
  \end{figure}
\begin{figure}[!tp]
    \centering
    \includegraphics[width=1.0\linewidth]{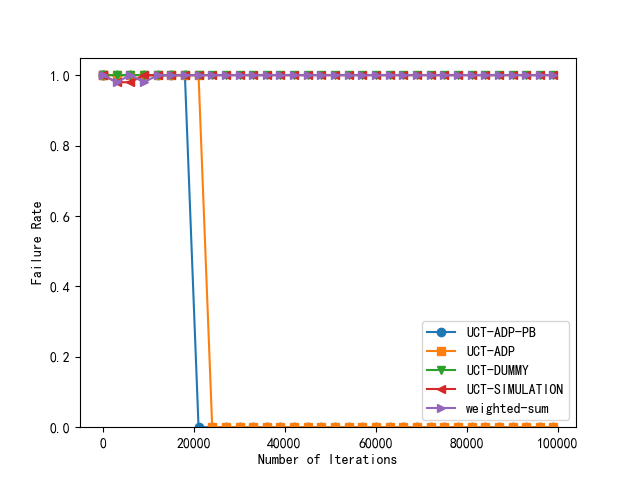}  
    \caption{Convergence to the correct move under Board Situation 2.}
    \label{convergence2}
  \end{figure}

  We also consider board situations where ADP alone fails to predict the best move. In Fig. \ref{board2}, the ADP predicts the red square to be the best move for the black side as it forms a double threat which leads to victory in most cases. In this specific case, however, the white side has a counterattack move (labeled with a green star). The counter move blocks the black four and forms a white live four at the same time, which leads to the defeat of black. The true best move for black that leads to victory is marked with the blue diamond. 
  \\
  \par
  Although ADP predicts the wrong move, it is still possible for UCT to correct the mistake by exploring down the game tree. As shown in Fig. \ref{convergence2}, the UCT-ADP deviate from the correct move in the first 20000 iterations due to ADP's failure, but converges to the right answer after 20000 iterations. The overall convergence speed is still faster than the other non-ADP baseline methods. The baseline methods do not converge even after 100000 iterations. 
\subsection{Computational Speed of UCT-ADP}
  We test for every model how many tree policy iterations can be achieved in a given time. The test is held under the board situation of Fig. \ref{board1}. It turns out that replacing the simulation process with ADP greatly improves the computational speed. When given 20 seconds, UCT-ADP and UCT-ADP-PB achieve over 20000 iterations while UCT-SIMULATION can only run 500 iterations due to the large simulation depth. Although UCT-DUMMY's iteration speed is twice of UCT-ADP (UCT-DUMMY does not use ADP to evaluate board situations and simply use a constant 0.5 for back-propagation, which reduces its computational cost), its convergence speed is unacceptably slow.
  \begin{figure}[!tp]
    \centering
    \includegraphics[width=1.0\linewidth]{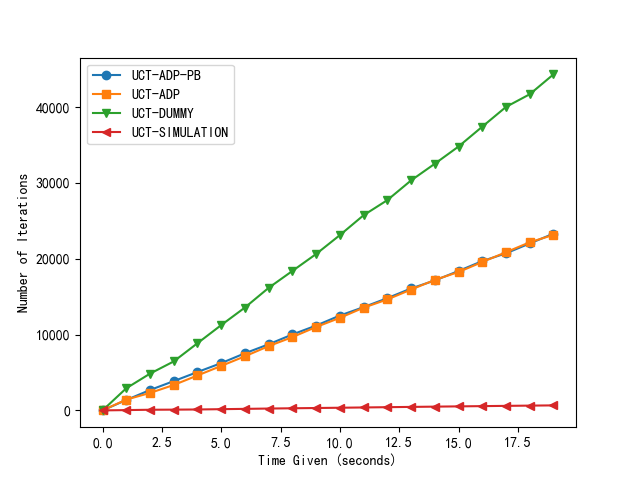}  
    \caption{Number of Iterations Achieved in a Given Time.}
  \end{figure}
\subsection{Playing Against Other Gomoku AIs}
  In order to evaluate the stability of our Gomoku AI, we let our agents play against other Gomoku agents in the Gomocup platform, including Mushroom, PureRocky, Valkyrie, pisq7. We also limit the game total time to 90 seconds and less than 15 seconds in one turn. Table \ref{play against} shows that the UCT-ADP Progressive Bias agent has better performance than both ADP and UCT after testing the efficacy of UCT-ADP Progressive Bias(UCT-ADP-PB) in gomocup's platform over 100 times.
\\
\par
\begin{table}[t]
  \caption{C\textsc{omparison} A\textsc{gainst} O\textsc{ther} A\textsc{gents}}
  \label{play against}
  \centering
  \begin{tabular}{lllll}
    \cmidrule{1-5}
    Agent     & MUSHROOM  & PureRocky  & Valkyrie  & Pisq7 \\
    \midrule
    ADP  &30:0  &27:3  &22:8  &10:20   \\
    UCT-SIMULATION  &15:15  &7:23  &3:27  &0:30  \\
    UCT-ADP  &30:0  &30:0  &27:3  &23: 7  \\
	UCT-ADP-PB  &30:0  &30:0  &28:2  &27:3  \\
    \bottomrule
  \end{tabular}
\end{table}

  Table \ref{play against}'s wining rate indicates ADP-UCT Progressive Bias's performance is more stable than both ADP’s and weighted sum ADP-UCT. Sometimes, it can also defeat some more powerful agent in gomocup. We are now trying to introduce the Memory-Augmented mechanism \cite{Memory-augmented monte carlo tree search} in this algorithm in order to enhance its situation sequence memory.
\\
\par
  The result implies that ADP and Progressive Bias are effective supports for UCT. ADP alone can make decent moves in a very short time (at around 100 milliseconds), while Progressive Bias improves the decisions made by UCT with the cost of time. UCT offers a structure to combine the advantage of both ADP and  Progressive Bias's heuristic knowledge.
\\
\par
Our code and Gomoku pattern description are open source in https://github.com/IrohXu/Gomoku-XYH19. The final version will continuously revise.
\\

\section{Conclusion and Outlook}
  Overall, we improve the method by employing ADP combined with UCT of MCTS algorithm to solve Gomoku in this paper. From the experiment, ADP and exponential heuristic function with MCTS is able to win most of the weak AIs in Gomocup in a time limit. However, it still has a certain gap with some stronger AI such as YiXin and Wine. One of the possible reason is that these powerful AI is written by C++ not Python, on the other hand, 32 patterns can not fulfill all complex situation on the Gomoku board. In the future, we will try to employ some simple pruning method such as MCTS-TSC in paper \cite{Monte-Carlo tree search with tree shape control} to decrease high branch factor and apply Memory-Augmented mechanism [13] to design and train a Memory-Augmented ADP.
\\


\section*{Acknowledgment}

This research project is supported by School of Data Science, Fudan University.
Also, We would like to thank Zhentao Tang and Dongbin Zhao whose idea inspire us to improve their Adaptive Dynamic Programming algorithm for Gomoku.
\\




\begin{thebibliography}{1}

\bibitem{A Survey of Monte Carlo Tree Search Methods}
C. B. {Browne}, E. {Powley}, D. {Whitehouse}, S. M. {Lucas}, P. I. {Cowling}, P. {Rohlfshagen}, S. {Tavener}, D. {Perez}, S. {Samothrakis}, and S. {Colton}, ``A Survey of Monte Carlo Tree Search Method,'' \emph{IEEE Transactions on Computational Intelligence and AI in Games}, vol. 4, no. 1, pp. 1-43, 2012.
\bibitem{Monte Carlo Tree Search Method for AI Games}
T. Patil, K. Amrutkar, and P. K. Deshmukh, ``Monte Carlo Tree Search Method for AI Games,'' \emph{The International Journal of Emerging Trends \& Technology in Computer Science}, vol. 2, 2013.
\bibitem{Effective Monte-Carlo tree search strategies for Gomoku AI}
J. H. {Kang} and H. J. {Kim}, ``Effective Monte-Carlo tree Search Strategies for Gomoku AI,'' \emph{International Journal of Circuit Theory and Applications}, vol. 10, pp. 4841-4843, 2016.
\bibitem{Evolving Gomoku solver by genetic algorithm}
{J. R. Wang}, {L. Huang}, and D. {Whitehouse}, ``Evolving Gomoku Solver by Genetic Algorithm,'' \emph{2014 IEEE Workshop on Advanced Research and Technology in Industry Applications}, 2014, pp. 1064-1067.
\bibitem{A.I for Games with High Branching Factor}
S. {Mohandas} and M. A. {Nizar}, ``A.I for Games with High Branching Factor,'' \emph {2018 International CET Conference on Control, Communication, and Computing}, 2018, pp. 372-376.
\bibitem{Monte-Carlo Tree Search and Rapid Action Value Estimation in Computer Go}
S. Gelly and D. Silver, ``Monte-Carlo Tree Search and Rapid Action Value Estimation in Computer Go,''  \emph{Artificial Intelligence}, vol. 175, no. 11, pp. 1856-1875, 2011.
\bibitem{Information Set Monte Carlo Tree Search}
P. I. {Cowling}, E. J. {Powley}, and D. {Whitehouse}, ``Information Set Monte Carlo Tree Search,'' \emph {IEEE Transactions on Computational Intelligence and AI in Games}, vol. 4, no. 2,  pp. 120-143, 2012.
\bibitem{Adaptive game AI for Gomoku}
{K. L. Tan}, C. H. {Tan}, K. C. {Tan}, and A. {Tay}, ``Adaptive Game AI for Gomoku,'' \emph {2009 4th International Conference on Autonomous Robots and Agents}, 2009, pp. 507-512.
\bibitem{Reinforcement Learning for Build-Order Production in StarCraft II}
Z. {Tang}, D. {Zhao}, Y. {Zhu}, and P. {Guo}, ``Reinforcement Learning for Build-Order Production in StarCraft II,'' \emph {2018 Eighth International Conference on Information Science and Technology}, 2018, pp. 153-158.
\bibitem{Self-teaching adaptive dynamic programming for Gomoku}
{D. Zhao, Z. Zhang, and Y. Dai}, ``Self-teaching Adaptive Dynamic Programming for Gomoku,'' \emph {Neurocomputing}, vol. 78, no. 1, pp. 23-29, 2012.
\bibitem{ADP with MCTS algorithm for Gomoku}
{Z. T. Tang}, D. {Zhao}, {K. Shao}, and {L. Lv}, ``ADP with MCTS Algorithm for Gomoku,'' \emph {2016 IEEE Symposium Series on Computational Intelligence}, 2016, pp. 1-7.
\bibitem{An algorithmic design and implementation of outer-open gomoku}
C. {Chen}, S. {Lin}, and Y. {Chen}, ``An Algorithmic Design and Implementation of Outer-open Gomoku,'' \emph{2017 2nd International Conference on Computer and Communication Systems}, 2017,  pp. 26-30.
\bibitem{Memory-augmented monte carlo tree search}
C. Xiao, J. Mei, and M. Muller, ``Memory-augmented Monte Carlo Tree Search,'' \emph{Thirty-Second AAAI Conference on Artificial Intelligence}, 2018.
\bibitem{Monte-Carlo tree search with tree shape control}
O. I. {Marchenko} and O. O. {Marchenko}, ``Monte-Carlo Tree Search with Tree Shape Control,'' \emph {2017 IEEE First Ukraine Conference on Electrical and Computer Engineering}, 2017,  pp. 812-817.
\bibitem{Bandit based monte-carlo planning}
L. Kocsis and C. Szepesvári, ``Bandit Based Monte-carlo Planning,'' \emph {ECML'06 Proceedings of the 17th European conference on Machine Learning}, 2006, pp. 282-293.

\end{thebibliography}
%

\end{document}